# CGEMs: A Metric Model for Automatic Code Generation using GPT-3

Aishwarya Narasimhan

*Student, M.Tech Electronics, B M S College of Engineering, aishwaryan.lel19@bmsce.ac.in*

Krishna Prasad Agara Venkatesha Rao

*Program Manager, Sony India Software Centre Pvt. Ltd.(SISCPL), avkrishna2003@gmail.com*

Veena M B

*Senior Member IEEE, Associate Professor, B M S College of Engineering, veenamb.ece@bmsce.ac.in*

**ABSTRACT**

Today, AI technology is showing its strengths in almost every industry and walks of life. From text generation, text summarization, chatbots, NLP is being used widely. One such paradigm is automatic code generation. An AI could be generating anything; hence the output space is unconstrained. A self-driving car is driven for 100 million miles to validate its safety, but tests cannot be written to monitor and cover an unconstrained space. One of the solutions to validate AI-generated content is to constrain the problem and convert it from abstract to realistic, and this can be accomplished by either validating the unconstrained algorithm using theoretical proofs or by using Monte-Carlo simulation methods. In this case, we use the latter approach to test/validate a statistically significant number of samples. This hypothesis of validating the AI-generated code is the main motive of this work and to know if AI-generated code is reliable, a metric model CGEMs is proposed. This is an extremely challenging task as programs can have different logic with different naming conventions, but the metrics must capture the structure and logic of the program. This is similar to the importance grammar carries in AI-based text generation, Q&A, translations, etc. The various metrics that are garnered in this work to support the evaluation of generated code are as follows: Compilation, NL description to logic conversion, number of edits needed, some of the commonly used static-code metrics and NLP metrics. These metrics are applied to 80 codes generated using OpenAI's GPT-3. Post which a Neural network is designed for binary classification (acceptable/not acceptable quality of the generated code). The inputs to this network are the values of the features obtained from the metrics. The model achieves a classification accuracy of 76.92% and an F1 score of 55.56%. XAI is augmented for model interpretability.

## 1 INTRODUCTION

In recent years, due to the rapid growth in AI technology, software engineering has advanced methods to directly generate codes conforming to a specific programming language just from Natural Language description. Code generation here does not refer to secret-key codes for encryption or QR codes but to programs(AI) that generate programs, which are usually coded by humans.

Earlier researches were mainly focused on intelligent “code completion”. Now many researchers are exploring the field of code summarization, code search, and code generation by natural language queries. Ling *et al*.[1] introduced a sequence-to-sequence model to generate code from natural language and structured specifications. Thanks to advancements in such technologies, there are various methods/tools for code generation in Python and Java, based on seq2seq models.

As industries are moving towards automation in every possible aspect, automatically generated codes are of great practical interest since they can help in software development in a lot many ways. The hypothesis under consideration is: “Using AI to generate code, reduces the time to market a software product”. This hypothesis is supported in this work considering bigger-picture questions like how to evaluate the code quality? What are the evaluation metrics to be considered? Which coding language to be considered? Can this code be reliable and robust?

Currently, though there are several researches regarding code generation, there aren't any good holistic researches carried out so far to validate the ***'AI-generated code'***, verified for its efficiency, correctness, and usability for the consequences described in Figure 1 [1]. The main utility of using AI is to reduce the burden and ease the completion of tasks. If automation or AI does not aid in this, the whole point of using AI becomes fruitless. These metrics serve to support the hypothesis, as well as a reference to future code generation techniques.

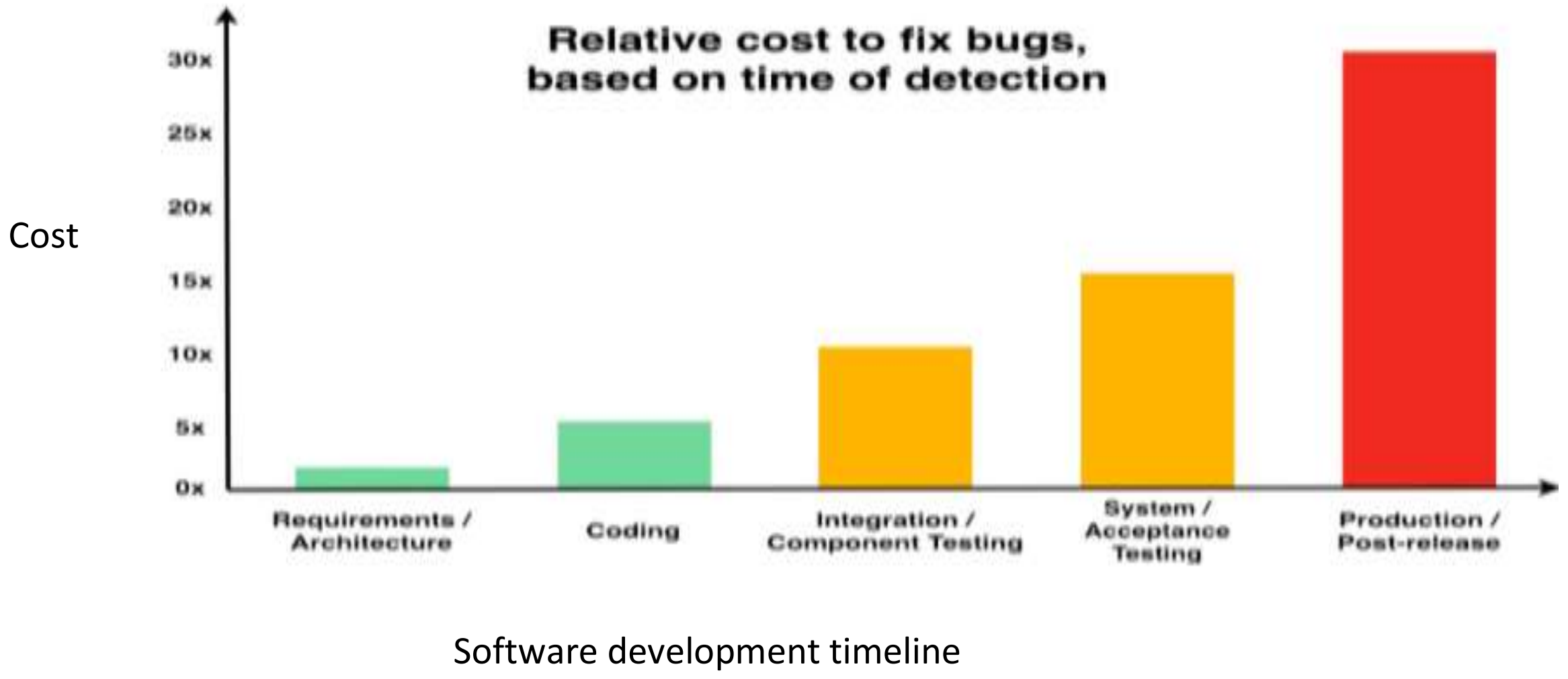

Figure 1: Relative cost of fixing bugs

## 2 RELATED WORK

Currently, there are several studies on static source code metrics. Zhen Li *et al.* [2] has given a brief survey of the existing literature as follows: "For the researches of the source code metrics model, Lin Y *et al.* [3] proposed a new metric model of JavaScript and React source code quality, which could be used to monitor source code quality for improving software development efficiency. Guigen L I [4] established a metrics model for the maintainability of object-oriented software, which included four sub-identity models of maintainability and a colligation model to the sub-identity. After discussing source code's reusability, Manoj H M [5] developed a metrics model of source code reusability. The relationship between design components of standard software metrics and source code reusability is established by using case studies from three software projects, which contained customer relationship management project, supply chain management project and enterprise relationship management project.

There are some studies of source code metric method. Based on the metric characteristics of Chidamber, Kemererand cyclomatic complexity, AlbertoS *et al.* [6] proposed the object-oriented metric methods after collecting the source code by systematic mapping. Núñez-Varela A S *et al.* [7] proposed a user-oriented tool to measure source code, in which the users can define their own metrics and incorporate new languages. Singh P *et al.*[8] proposed a software quality assurance tool for measuring source code of C# at the class and method levels."

Alexey Svyatkovskiy *et al.*[9] propose an end-to-end code sequence completion system called IntelliCode Compose based on the GPT-C (variant of GPT-2).They use perplexity to evaluate the quality of language model pretraining of GPT-C models and ROUGE-L and the Levenshtein similarity as the evaluation metrics to measure offline performance of the code sequence completion system. Wang proposes an Autocoder [10] by finetuning GPT-2 for Auto Code Completion in Python and Java. Wang concludes that 'although some generated examples look good, it needs to take a grain of salt to judge the model's actual performance. The model may simply **remember** existing code in the training set well'. Besides, Galois[2] is an auto code completer for code editors

[1] https://deepsource.io/blog/exponential-cost-of-fixing-bugs/

[2] https://github.independentlyreview.com/galois-autocompleter/galois-autocompleter

(or any text editor) based on OpenAI GPT-2 and Tabnine[3] also uses GPT2 to serve ranked lists of code sequence suggestions.

However, there are very few researches that focus completely on evaluating the AI-generated code which is presented next. Zhen Li *et al*.[2] proposed a metric model of automatic code generation based on the quality and efficiency of the Java and Python codes generated. CodeGeneration, java-codetool and aiXcoder were selected as the tools of automatic code generation for their experiments. Colin B Clement *et al.* [11] presented PyMT5 which is a text-to-text transformer model for method generation and docstring generation and compared their method with GPT-2 using BLEU, ROUGE, Perplexity, and Syntax. There aren't any other researches that holistically look at validating the AI-generated code.

## 3 PROPOSED METHOD – CGEMS MODEL

This work is carried following the CRISP-DM Methodology. As detailed in section 2, there aren't many researches which holistically look at validating the AI-generated code. Hence a metric model called CGEMs (Code Generation Evaluation Metrics) is proposed, which consists of the following metrics:

- **Compilation**

This metric relates to whether the code generated is compiling or not.

- **Functionality**

The code that is generated based on natural language description must match the functionality desired.

- **Number of compilation errors**

How many compilation errors are present? However, in the case of interpreter-based languages like Python, all the compilation errors are not listed once unlike in C, C++.

- **Number of edits required**

How many edits are required to make the code compile and/or the functionality correct? Each addition, deletion, and modification are considered as one edit/statement.

- **Sequence ratio**

SequenceMatcher is a class in the standard Python library 'difflib' which compares 2 documents (here 2 codes) and gives a normalized "ratio of similarity" in the range 0-1. The formula used is

$$\text{Sequence ratio} = 2.0*M / T$$

where T is the total number of elements in both sequences, and M is the number of matches.

The same difflib library is used to highlight the differences between generated code and corresponding corrected code and hence find the number of edits.

- **Execution time**

Execution time is measured in microseconds to understand the time complexity.

- **Code coverage**

Code coverage is performed to verify the extent to which the code has been executed. It is a commonly used metric at unit-level testing by the developers to understand which test cases need to be added, removed, or modified.

- **Code similarity**

[3] https://tabnine.com/blog/deep/

To measure code similarity, cosine and soft cosine similarity are used. Cosine similarity measures the cosine of the angle between two vectors projected in a multi-dimensional space. In this context, the two vectors are arrays containing the word counts of two codes. The soft-cosine similarity is a metric used to compare the similarity of 2 documents while also considering the semantic meaning. Smaller the angle, similar the vectors x and y.

If x and y are row vectors as in Figure 2, their cosine similarity k is defined as:

$$k(x, y) = \frac{xy^{\top}}{\| x \| \| y \|}$$

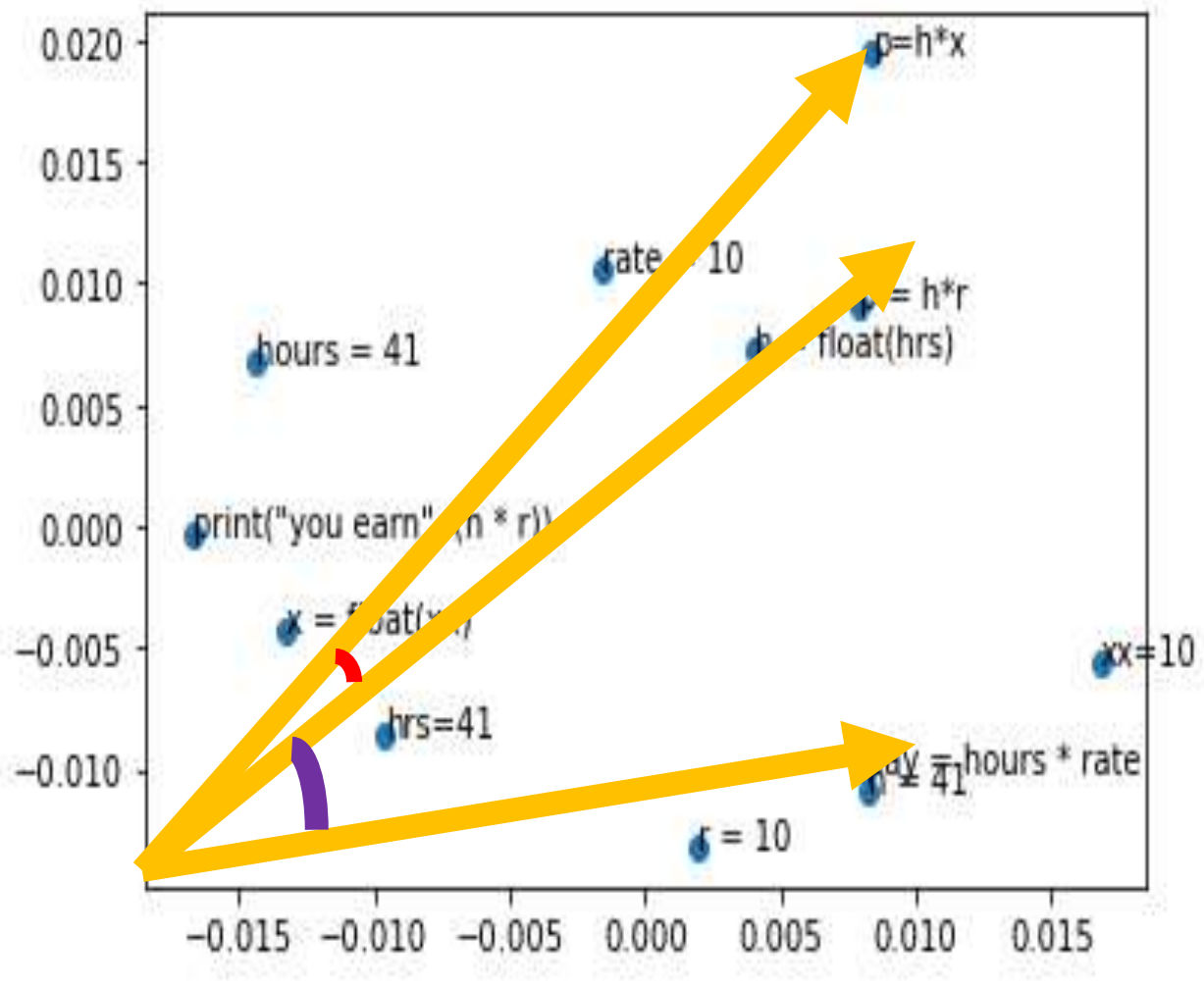

Figure 2: Glimpse of Cosine similarity for the code to calculate pay, given rate/hour and the number of hours by 3 different developers.

- **BLEU score**

BLEU (BiLingual Evaluation Understudy) is a metric that is originally used for evaluating machine-translated text.

- **ROUGE- 1, 2, L score**

ROUGE (Recall-Oriented Understudy for Gisting Evaluation) is a metric originally used to evaluate automatic summarization of texts.

- **Code quality – maintainability**
- **Code quality – complexity**
- **Code quality – Halstead**
- **Code quality - Raw metrics**

Static analyzers like Klocwork are used by developers to ensure that the code is of high quality. Developers will be able to monitor the quality of the codebase over time, using metrics like cyclomatic complexity. Maintainability index etc.

Raw metrics include:
LOC (Lines of code): Total number of lines of code.
LLOC (Logical Lines of Code): The number of logical lines of code.
Comments: Indicate how well the code is readable.

Cyclomatic complexity is a measure of how many linearly independent code paths there are through the code.
Halstead complexity measures the code complexity by relating to the size of a program's codebase.
Maintainability Index is a software metric that measures how easy it is to change the source code.

Metrics like Cosine Similarity, ROUGE, BLEU considered, need a reference document to provide a score. To provide a reference, student developers were asked to write codes by conducting an online quiz. The human-written codes were used as a golden reference against which the AI-generated codes were compared to obtain these metrics. ROUGE, BLEU score and code similarity capture string similarity of generated code and the reference code.

All the existing research[11] highlight these 3 metrics as major parameters to judge the quality of AI-generated code. But these metrics have a major drawback. This is justified based on the results shown in Figure 3.

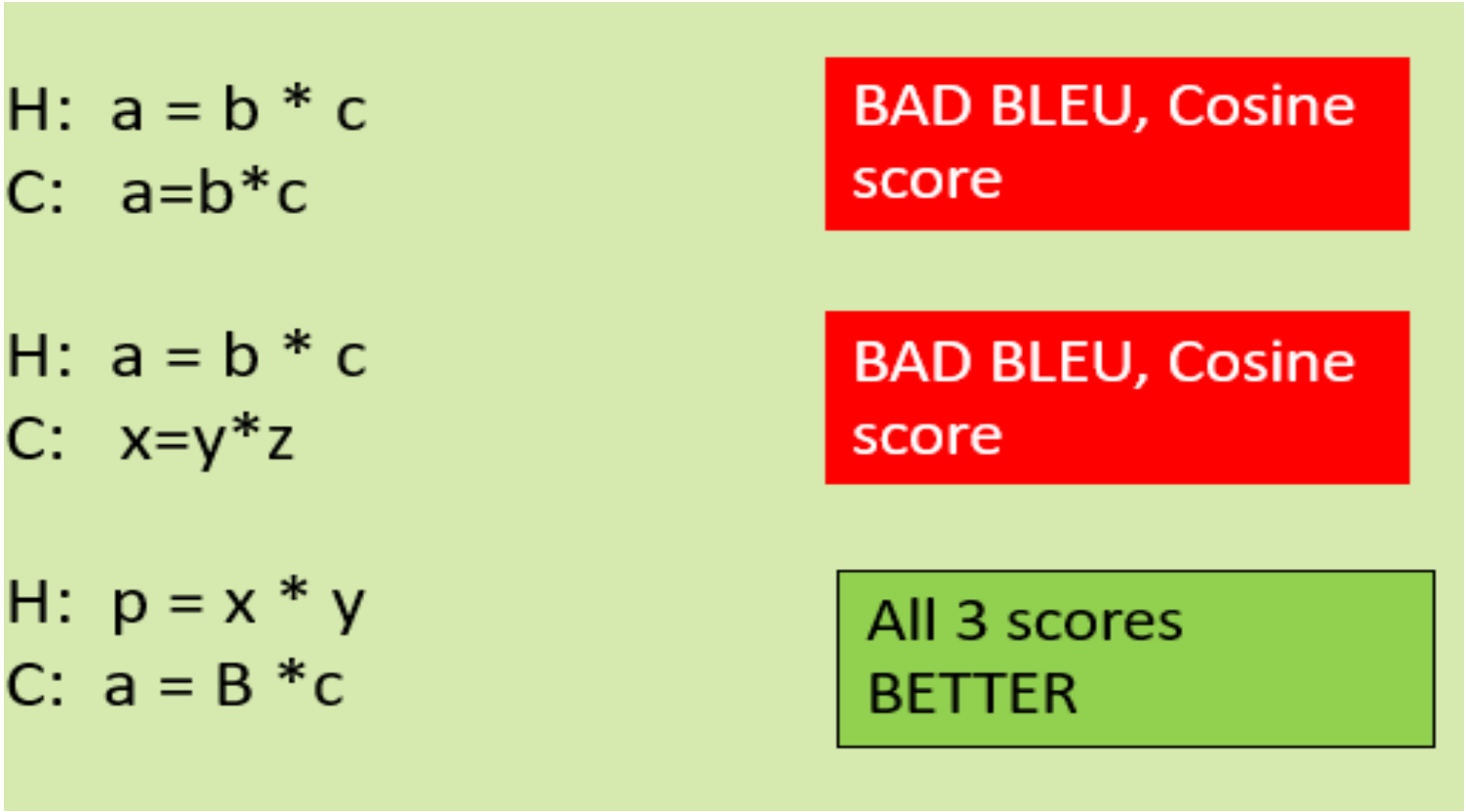

Figure 3: Performance of cosine similarity, BLEU, ROUGE to find the product of 2 numbers in Python. (H represents the hypothesis and C represents the reference/candidate)

As can be readily perceived from Figure 3, all the 3 codes correspond to multiplication of 2 numbers. Though the logic is identical, BLEU, ROUGE and cosine scores give better results for case-3 alone because of the spacing and other parameters which are not related to the logic of the code. Hence these parameters alone are not ideal metrics to measure the similarity between two codebases. Albeit these are good enough parameters for measuring the similarity between two textual documents, which contain grammar and syntax, that are heavily dependent on word sizes, spaces and capitalization.

### 3.1 Artificial Neural Network Modelling

An ANN was designed by providing input to the network as follows. Eighty AI-generated programs were executed and the above-mentioned metrics were captured for each program into a CSV file. This resulted in a total of 30 metric which is fed as input to the ANN.

The main reason for using NN model is to decipher how these 30 metric values (features) interplayed amongst themselves to decide the efficacy of the generated code. The hidden pattern created due to the interplay of features is very hard for humans to capture, as they are not cognizable to a human mind. The best way to highlight these hidden patterns is by training an ANN that can decipher the subtle variations of the features necessary in determining the quality of the generated code.

An ANN is designed to do a binary classification with two classes: Class-1 means the AI-generated code is reliable, good, or acceptable quality and Class-0 means AI-generated code is of bad quality i.e., usage of AI did not ease software development. A supervised learning mechanism is used and the ground truth is generated manually. The conditions for a generated program to be considered as class-1 is as below:

(i) The code needs to compile.
(ii) Partial or complete functionality is achieved.
(iii) Valid comments are present.
(iv) Any code which does not satisfy the above three conditions but can turn into a code that satisfies these conditions with less than or equal to 3 edits, then such programs are considered as class-1.

Rest all programs which do not fall under these conditions are considered class-0.

Table1 gives the attribute description. Among them, Code Complexity Grade is not considered as input since codes having Code Complexity numbers 1-5 have the same Code Complexity grade A. Also, Code Complexity is computed by Radon library only for those codes that have a function, Method, or Class.

Classification performance – Confusion matrix

Actual

Predicted

| | 1 | 0 |
|---|---|---|
| 1 | TP | FP |
| 0 | FN | TN |

Accuracy measures the correct predictions out of total predictions to be made.

*Correct predictions = TP + TN*

*Accuracy = Correct Predictions / Total Predictions*

- Precision summarizes the fraction of examples assigned to the positive class that belongs to the positive class.

***Precision*** *= True Positive / (True Positive + False Positive)*

- Recall summarizes how well the positive class was predicted.

***Recall*** *= True Positive / (True Positive + False Negative)*

- The Harmonic mean of Precision and recall is called the F-score or the F-measure.

***F-Measure*** *= (2 * Precision * Recall) / (Precision + Recall)*

## 4 CODE GENERATION TOOLS

There are no open-source AI generating codes/tools that were available to generate codes and evaluate the metrics. So, the work was started by creating own datasets i.e., 65 Python programs for 20 problems: 2-5 solutions per problem (solutions taken from GeeksforGeeks[4], Stackoverflow[5], Programiz[6], w3resource[7] etc. are logically different and/ or way of writing code is different). Later as beta-access to GPT-3 [8] was obtained, 80 Python programs were generated with the help of Epsilon-code [9] by giving description in Natural Language.

## 5 EXPERIMENTS AND RESULTS

Various Python libraries and readily available packages were used to calculate metric values.

- The below command is used to check whether the code is compiling or not.

```
compile(source,filename,mode,PyCF_ONLY_AST)
```

- Difflib is used to find the sequence ratio[10]
- Coverage is used to find the statement coverage [11]
- Radon for the code quality metrics[12]
- Scikit-learn for cosine similarity [12]
- Genism for soft cosine similarity [13]
- NLTK for BLEU [14]

[4] https://www.geeksforgeeks.org/python-programming-language/
[5] https://stackoverflow.com/
[6] https://www.programiz.com/python-programming/
[7] https://www.w3resource.com/python-exercises/
[8] https://beta.openai.com/
[9] https://epsilon.shreenabh.com/
[10] https://docs.python.org/3/library/difflib.html
[11] https://coverage.readthedocs.io/en/coverage-5.5/
[12] https://radon.readthedocs.io/en/latest/

- PyRouge from rogue-metric for ROUGE scores[13]

For the codes which needed no correction, the sequence ratio is taken as 1 and the # of edits as 0

Table 1: Description of each attribute

| Feature | Type | Description |
|---|---|---|
| Program | object | metadata |
| Code Coverage | int64 | % out of 100 |
| Maintainability Index | flaot64 | Out of 100 |
| Compiling | int64 | 1 = True, 0 = False |
| Functionality | int64 | 2 = Yes,1 = partial, 0 = No |
| Edits | int64 | 0,1,2… (discrete) |
| Sequence Ratio | float64 | Range b/w 0.0 – 1.0 |
| CC Grade | object | A B C D E |
| CC Number | float64 | 1-5,6-10,11-20,21-30,31-40,41+ |
| LOC | int64 | # of Lines of code (discrete) |
| ROUGE -1, 2, L | float64 | ROUGE – 1gram, 2gram, longest-matching sequence: Precision, Recall, F1 (out of 100) |
| LLOC | int64 | #of Logical Lines |
| SLOC | int64 | # of Source Lines of code (discrete) |
| Comments | int64 | 0,1,2… |
| C%L | int64 | % out of 100 |
| C%S | int64 | % out of 100 |
| C+ M % L | int64 | % out of 100 |
| Difficulty | float64 | Halstead value |
| Effort | float64 | Halstead value |
| Programming Time | float64 | Time required to program in seconds |
| Bugs | float64 | Number of bugs delivered |
| Execution Time | float64 | in microseconds |
| Cosine similarity | float64 | Degree (0 – 90) |
| Soft Cosine similarity | float64 | Degree (0 – 90) |
| BLEU | float64 | BLEU score (out of 100) |

Out of eighty GPT-3 generated programs, 42 programs have target label as class-0 and 38 as class-1. SMOTE Technique is used to improve efficacy. Thus, the dataset has 84 samples * 30 metric-values/sample with 42 samples for each class. 80% of the correlated features were removed, depicted in Figure 4. So instead of 30 values, only 15 values are sent to ANN , referred to as Model-1

**Input layer:** 15 metric values randomized and applied as input to 15 neurons.

**Training:** 71 samples

[13] https://pypi.org/project/rouge-metric/

**Testing:** Remaining 13 samples

**Hidden layers:** 14 and 12 neurons in 2 hidden layers respectively with ReLU activation.
**Output layer:** Softmax activation function with 2 neurons, Adam optimizer, Learning Rate = 0.001, Loss = Sparse Categorical cross-entropy, Epochs =1000

Binary classification accuracy: 76.81% during training and 61.54% during testing.

F1 metric: 61.87% during training and 55.56% during testing.

*Predicted :* *[**1**, 1, 0, **1**, 1, 0, 1, 1, 0, 0, **1, 0, 1**]*

*Actual label:* *[**0**, 1, 0, **0**, 1, 0, 1, 1, 0, 0, **0, 1 ,0**]*

The number of misclassifications = 5/13

To understand and verify what features did the NN looked into while making a prediction, the NN model is combined with XAI (Explainable AI) in Figure 5. ML Interpretable algorithm called LIME (Local Interpretable Model-agnostic Explanations) for tabular data is used for this purpose[15].

Further, 15 features are passed to SelectKBest API of Sci-kit learn. The p-values and the ANOVA F-value score are also shown in Figure 4(b).

From Figure 4(b), observe that not all 15 metrics have a major contribution. Among them, only 8 are selected, which are highlighted in green. Based on the number of features, a modified ANN model, referred to as Model -2 which has only 8 neurons in the input layer is designed. The remaining architecture and training mechanism is the same as that of Model-1 described above. The accuracy increased to 76.92% on test samples as in Figure 6.

The # of misclassifications = 3/13

*Predicted :* *[0, 1, 0, **1**, 1, 0 , 1, **0**, 0, 0, 0, **0**, 0]*

*Actual label:* *[0, 1, 0, **0**, 1, 0, 1, **1**, 0, 0, 0, **1** ,0]*

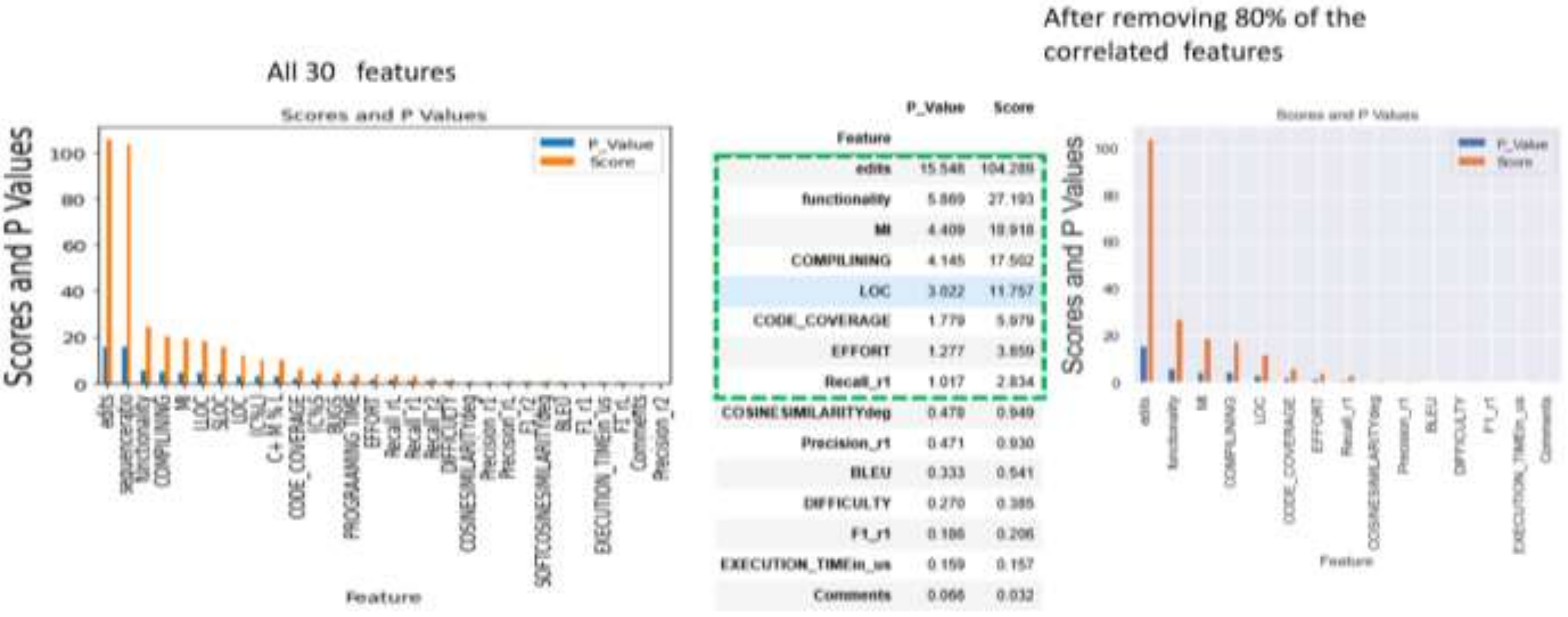

(a) (b) (c)

Figure 4: (a) All 30 Features (b) p-value and ANOVA F-scores for 15 features (c) Features considered for Model-1, after removing 80% of the correlated features among (a)

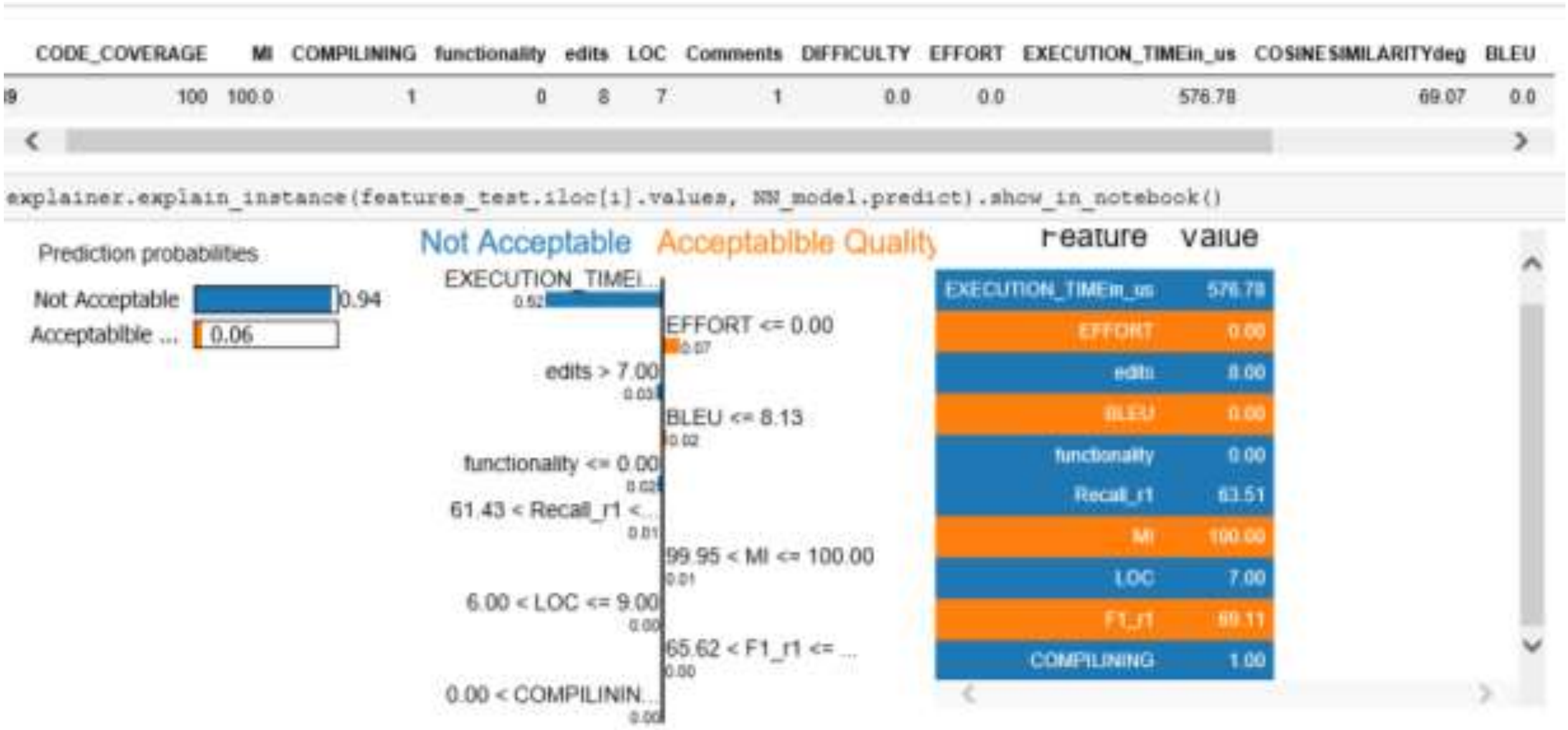

Figure 5: LIME output for a test sample

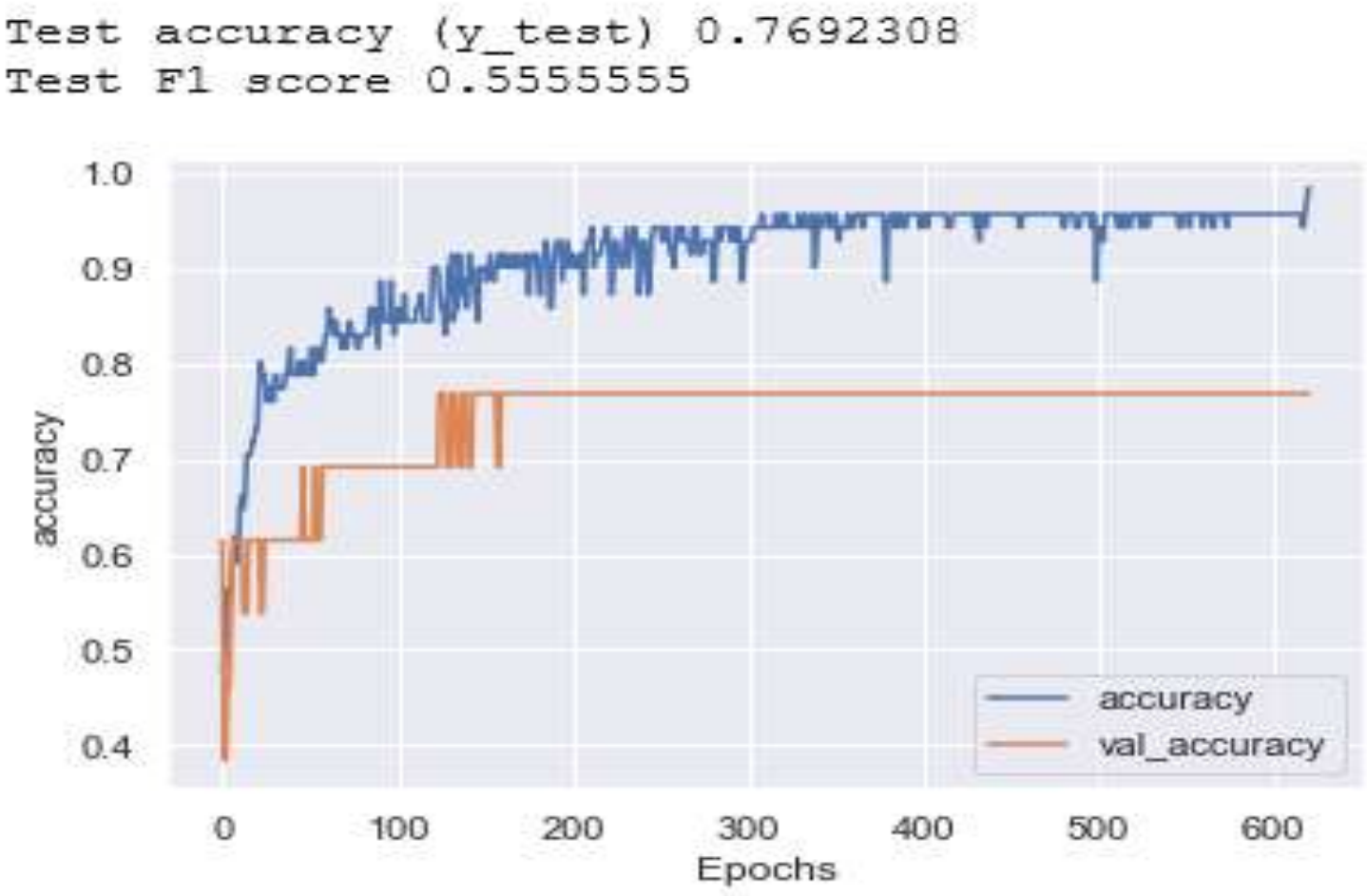

Figure 6: Accuracy during training and testing of Model-2

## 6 CONCLUSION AND FUTURE WORK

The metric model CGEMs for automatic code generation was studied in this work on a limited number of AI-generated codes. To the best of our knowledge, this is the first work in which ANN has been combined with code-metrics to assess the AI-generated codes. To the best of our knowledge, this is also the first work to evaluate the code generated by GPT-3. The performance of the two ANN models is summarized in Table 2.

Table 2: Summary of the ANN model performance

| Parameters | Model-1 | Model-2 |
|---|---|---|
| Number of features | 15 | 8 |
| Accuracy on test samples | 61.54 % | 76.92 % |

Thus, we prove that our hypothesis is true. Artificially generated code can help in software development partially, if not completely. Despite these results, there are still some hindrances to evaluate only the logic of the code which can be taken up as future work.

For future work, further research could be considered in the following angles:

(i) For code similarity, code2vec as proposed by Uri Alon *et al.* [16] can be used to find the dot product of the vectors that represent the codes and an angular-threshold can be determined using Logistic regression to know the range of angles for which the programs are exactly same, similar and dissimilar.
(ii) On similar lines, cosine similarity can be used.

But the above two approaches would require a large number of code samples. Hence these experiments were not conducted as part of this work.

(iii) Abstract Syntax Tree (AST) can be utilized to understand the logical complexity by examining the structure of the source code at a deeper level.

## ACKNOWLEDGEMENT

This work was carried during the Internship at Sony India Software Centre Pvt. Ltd., Bangalore. The author[1] would like to thank BMS College of Engineering and Sony India Software Centre Pvt. Ltd. for the Internship opportunity. We would also like to express our gratitude to OpenAI's GPT-3 team and Epsilon for the beta access and assistance to generate codes, also to the student developers who participated in the quiz.